\documentclass{article}

\usepackage{microtype}
\usepackage{amsmath}
\usepackage{amsfonts}
\usepackage{amssymb}
\usepackage{fancyvrb}
\usepackage{graphicx}
\usepackage{subfigure}
\usepackage{booktabs} 

\usepackage{hyperref}

\usepackage[accepted]{icml2020}

\icmltitlerunning{Adaptive Modeling Against Adversarial Attacks}

\begin{document}

\twocolumn[
\icmltitle{Adaptive Modeling Against Adversarial Attacks}



\icmlsetsymbol{equal}{*}

\begin{icmlauthorlist}
\icmlauthor{Zhiwen Yan}{nus}
\icmlauthor{Teck Khim Ng}{nus}
\end{icmlauthorlist}

\icmlaffiliation{nus}{School of Computing, University of Singapore, Singapore}

\icmlcorrespondingauthor{Teck Khim Ng}{ngtk@nus.edu.sg}

\icmlkeywords{Machine Learning, ICML}

\vskip 0.3in
]



\printAffiliationsAndNotice{\icmlEqualContribution} 

\begin{abstract}
Adversarial training, the process of training a deep learning model with adversarial data, is one of the most successful adversarial defense methods for deep learning models. We have found that the robustness to white-box attack of an adversarially trained model can be further improved if we fine tune this model in inference stage to adapt to the adversarial input, with the extra information in it. We introduce an algorithm that “post trains” the model at inference stage between the original output class and a “neighbor” class, with existing training data. The accuracy of pre-trained Fast-FGSM CIFAR10 classifier base model against white-box projected gradient attack (PGD) can be significantly improved from 46.8\% to 64.5\% with our algorithm. 
\end{abstract}

\section{Introduction}

Although deep learning models have been very successful in many applications, its vulnerability to adversarial input is a significant concern \cite{szegedy2014intriguing}.  Small perturbations to a clean input could potentially mislead the model to produce very different output. Many researchers have introduced defense algorithms that included adversarial training \cite{madry2019deep}, randomization \cite{xie2018mitigating}, and denoising \cite{xu2018feature}.  Adversarial training is one of the most successful defense algorithms compared to many other algorithms that were later broken after stricter evaluation \cite{athalye2018obfuscated}.

\begin{figure*}[ht]
\begin{center}
\centerline{\includegraphics[width=1.5\columnwidth]{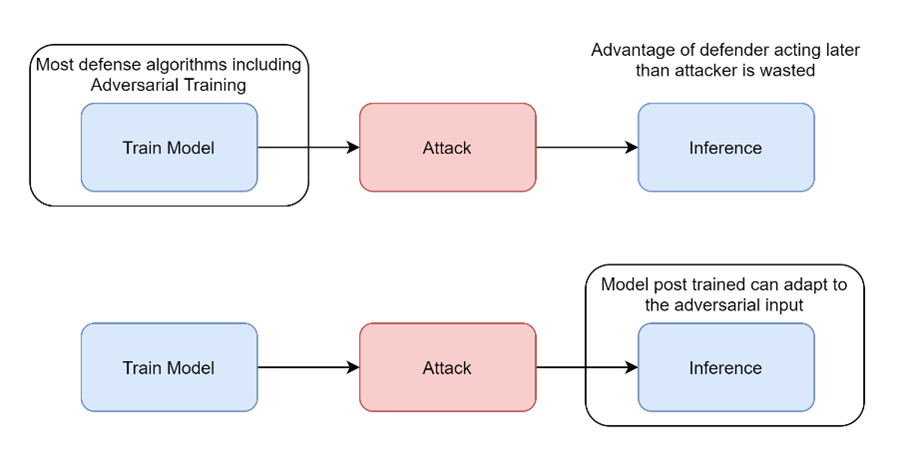}}
\caption{Most defense algorithms aim to produce robust but fixed networks at training stage, which gives the attacker an information advantage. We propose to post train models after attacks at inference stage, which gives the defender the advantage of knowing the adversarial input and adapt to it.}
\label{fig:pt_illustration}
\end{center}
\vskip -0.2in
\end{figure*}

Adversarial training trains the model with adversarially generated input instead of clean input, but the model is fixed once the training process has ended. In the white-box attack settings, the attacker has perfect information of this model, including its structure and parameters. Gradient based attackers, Fast Gradient Sign Method (FGSM) \cite{goodfellow2015explaining} for example, can then accurately calculate the gradient of the model and produce accurate adversarial input. It is obvious that this perfect information gives the defender significant disadvantage because the trained model is fixed and not adaptive to adversarial input. The attack and defense scenario is asymmetric as the attacker has full information of the defender, while the defender could not make use of knowledge of the attacker. Intuitively, the attacker attacks with the knowledge of the model, we should equip the defender with knowledge of any particular adversarial input, as shown in \textbf{Figure} \ref{fig:pt_illustration}. A defender should adapt to any particular input before inferring it. 

To achieve better robustness of the model, we propose to “post train” the model with small amount of data from only two classes induced by particular input, the original class and “neighbor” class. The original class is the prediction of the submitted input, and the neighbor class is the prediction of the input after a self-induced round of adversarial attack. We argue that the correct prediction will most likely fall within these two classes. The model can then focus on deciding which these two classes is correct. 

Our technique can be used to strengthen any adversarially trained models. We apply our technique to existing adversarially trained models and compare their robustness before and after our modification. We evaluate the accuracy of the models using both the standard adversarial training  (Madry Model) \cite{madry2019deep} and efficient adversarial training  (Fast-FGSM model) \cite{wong2020fast}, under white-box untargeted projected gradient attacks (PGD) \cite{madry2019deep} on CIFAR10 dataset \cite{Krizhevsky2009LearningML} and MNIST dataset \cite{mnist}. In CIFAR10 dataset, we significantly increased the robust accuracy of the Fast-FGSM base model from 46.8\% to 64.5\% and the robust accuracy of the Madry base model from 47.8\% to 54.9\%, under $l_{\infty}$ PGD attack. In MNIST dataset, we slightly improve the robust accuracy of the Fast-FGSM base model from 92.0\% to 94.4\% and the robust accuracy of Madry base model from 95.4\% to 96.2\%, under $l_{\infty}$ PGD attack. 

\section{Related Works}
Vulnerability of deep learning models against adversarial perturbation was first discovered in \cite{szegedy2014intriguing}, Fast Gradient Sign Method (FGSM) \cite{goodfellow2015explaining} was proposed to efficiently generate untargeted adversarial input using first order information of a model. FGSM was later improved to Projected Gradient Descent (PGD) \cite{madry2019deep} that performed FGSM in multiple steps to find a better adversarial input. Although many other attack algorithms were introduced later, for example, Carini and Wagner (C\&W) \cite{carlini2017evaluating} and  DeepFool \cite{moosavidezfooli2016deepfool}, PGD is arguably the best universal first-order adversary \cite{madry2019deep}.  Models robust to PGD should be robust to all adversarial attacks relying on first-order information of the model. PGD is also widely used in evaluating the robustness of defensive models because of this reason.

Many defensive algorithms were also introduced against adversarial attack, including data compression \cite{dziugaite2016study}, data randomization \cite{wang2017learning}, regularization \cite{zhang2019theoretically} and defensive distillation \cite{papernot2016distillation}. Although some of these algorithms were later broken under stronger and iterative attack methods, adversarial training remains one of the most effective defensive methods until today. Adversarial training was first proposed \cite{madry2019deep} to solve the inner optimization problem of defense together with the outer optimization problem of attack. Many other variants of adversarial training were also introduced later, for example, adversarial logit pairing \cite{kannan2018adversarial}, generative adversarial training \cite{lee2017generative}, adversarial training for multiple types of attacks \cite{tramer2019adversarial} and adversarial training using unlabeled data \cite{uesato2019labels} . One important drawback of adversarial training is its slow training speed due to the repeated steps in inner optimization process. Therefore efficient adversarial training algorithms were also introduced to address this issue including free adversarial training \cite{shafahi2019adversarial} and fast adversarial training \cite{wong2020fast}.

\section{Algorithm Design}
\subsection{Adaptive Models at Inference Stage}
Most of the adversarial training algorithms  are based on a unified view of attacks and defense during the training phase. They all try to solve the following optimization problem:

\begin{align}
\label{eqn:adv_train}
\min _{\theta} \mathbb{E}_{(x, y) \sim D}\left\{\max _{\delta \in \Delta} \mathcal{L}\left(f_{\theta}(x+\delta), y\right)\right\}
\end{align}

where $\mathcal{L}$ is the loss function, $f_\theta$ is the model parameterized by $\theta$, D is the training distribution of $\left(x,y\right)$ pairs. This optimization problem consists of an inner maximization problem which represents the adversarial attack, and an outer minimization problem which represents the adversarial defense. 

To solve this saddle point optimization problem, adversarial training algorithms generate adversarial inputs during the training phase and aim to minimize the loss produced by the model. However, once the training phase has ended, the same model will be used during the inference phase passively accepting the adversarial inputs. 

In inference stage, attackers solve the inner maximization problems, but the defender could not further optimize the model after receiving this adversarial input.

In white-box attack scenario, the attacker has the full information of this model. This gives the defender a significant disadvantage. If we allow the defending model to adapt to each adversarial input, it will receive the following advantages: a) The final model will have stochastic nature in the eyes of the attacker even with the full information of the model before the inference stage, making the attack harder. b) The final model can focus on minimizing the risk of this particular input instead of the population risk, making the defense more targeted and therefore more effective. This algorithm of fine tuning the model at inference stage based on the information derived from the adversarial input is named “post training” for easy reference.

\subsection{Finding Original Class and Neighbor Class}
\label {post_train_diff_algo}
To post train the model at inference stage, with the knowledge of adversarial input $x^\prime=x+\delta$, we have a few possible options:

\begin{itemize}
\item Optimize directly on $\min _{\theta}\left\{\mathcal{L}\left(f_{\theta}\left(x^{\prime}\right), y\right)\right\}$, but this is difficult because the defender does not know the ground truth label y

\item 	Optimize on the entire training dataset to solve $\min _{\theta} \mathbb{E}_{(x, y) \sim D}\left\{\max _{\delta \in \Delta} \mathcal{L}\left(f_{\theta}(x+\delta), y\right)\right\}$ which is the same as training stage. This gives some stochasticity to the model, but does not help the model to focus on the known adversarial input $x^\prime$

\item 	Optimize on data with two different labels, $y^\prime$ and $y^{\prime\prime}$, with either one of them having high probability to be the ground truth label of the input. This means $y^\prime$ and $y^{\prime\prime}$ are chosen such that $P\left(y=y^\prime\vee y=y^{\prime\prime}\right)$ is as large as possible. 
\end{itemize}

We propose to post train with option (c), by finding the two highly probable class labels of the adversarial input in the following manner:

\begin{enumerate}
\item We define $y^\prime$ as the original class, which is just the output class of the adversarial input $x^\prime$ from our trained model $f_\theta$:
\begin{align}
y^\prime=f_\theta\left(x^\prime\right)
\end{align}

\item 	We define $y^{\prime\prime}$ as the neighbor class, which is the output class of another untargeted adversarial input $x^{\prime\prime}$, based on the adversarial input $x^\prime$ and original class $y^\prime$:
\begin{align}
\begin{gathered}
x^{\prime \prime}=x^{\prime}+\underset{\delta^{\prime} \in \Delta}{\operatorname{argmax}} \mathcal{L}\left(f_{\theta}\left(x^{\prime}+\delta^{\prime}\right), y^{\prime}\right) \\
y^{\prime \prime}=f_{\theta}\left(x^{\prime \prime}\right)
\end{gathered}
\end{align}

\end{enumerate}

We claim that in most cases, one and only one of these two classes $y^\prime$ and $y^{\prime\prime}$ are highly probable candidates for the ground truth.  In other words, the value of $P\left(y=y^\prime\vee y=y^{\prime\prime}\right)$ is relatively large. We will support this claim with intuitions in boundaries and actual image classification experimental results in \textbf{Section} \ref{neighbor_acc_exp}. The detailed algorithm of finding the neighbor class is described in \textbf{Algorithm} \ref{alg:neighbor}.

\begin{algorithm}[tb]
   \caption{Find neighbor class}
   \label{alg:neighbor}
\begin{algorithmic}
   \STATE {\bfseries Input:} Robust model $f_\theta$ with parameter $\theta$, adversarial input $x^\prime$, PGD attack iteration $L$ ($L$=20 in case of $PGD^{20}$)
   \STATE {\bfseries Output:} Neighbor class $y^{\prime\prime}$
   \STATE $\delta^\prime=0$
   \FOR{$i=1$ {\bfseries to} $L$}
   \STATE $\delta^{\prime} = \delta^{\prime}+\alpha \cdot \operatorname{sign}\left(\nabla_{\delta^{\prime}} \mathcal{L}\left(f_{\theta}\left(x^{\prime}+\delta^{\prime}\right), y^{\prime}\right)\right)$
   \STATE $\delta^{\prime}=\max \left(\min \left(\delta^{\prime}, \epsilon\right),-\epsilon\right)$
   \ENDFOR 
   \STATE $x^{\prime \prime}=\max \left(\min \left(x^{\prime}+\delta^{\prime}, 1\right), 0\right)$
   \STATE Neighbor class $y^{\prime\prime}=f_\theta (x^{\prime\prime})$
\end{algorithmic}
\end{algorithm}

\subsection{Why Ground Truth Label is Between Original Class and Neighbor Class}

To understand why the ground truth label $y$ has a high probability to be between the original class $y^\prime$ and neighbor class $y^{\prime\prime}$, or in other words,$P\left(y=y^\prime\vee y=y^{\prime\prime}\right)$ is high,  we need to discuss three different cases:

\begin{enumerate}
\item The input is natural without adversarial attack
\item The input is adversarially generated, but it is not a successful attack
\item The input is adversarially generated and it is a successful attack
\end{enumerate}

The first two cases are trivial, because no attack and unsuccessful attack implies $y=y^\prime$. Hence a model correctly classifying a natural input will have $P\left(y=y^\prime\vee y=y^{\prime\prime}\right)=1$ in these two cases.

The third case requires us to find the neighbor class $y^{\prime\prime}=y$ with high probability because $y\neq y^\prime$. We propose doing this by performing another untargeted attack on the adversarial input $x^\prime$ and output $y^\prime$. What an untargeted attack algorithm $x^{\prime\prime}=Attack\left(x^\prime,\ y^\prime\right)$ does is to find another input $x^{\prime\prime}$ in the neighborhood (hypercube of size $2\epsilon$) around the current input $x^\prime$, with a different class output other than $y^\prime$.

\begin{figure}[ht]
\begin{center}
\centerline{\includegraphics[width=0.7\columnwidth]{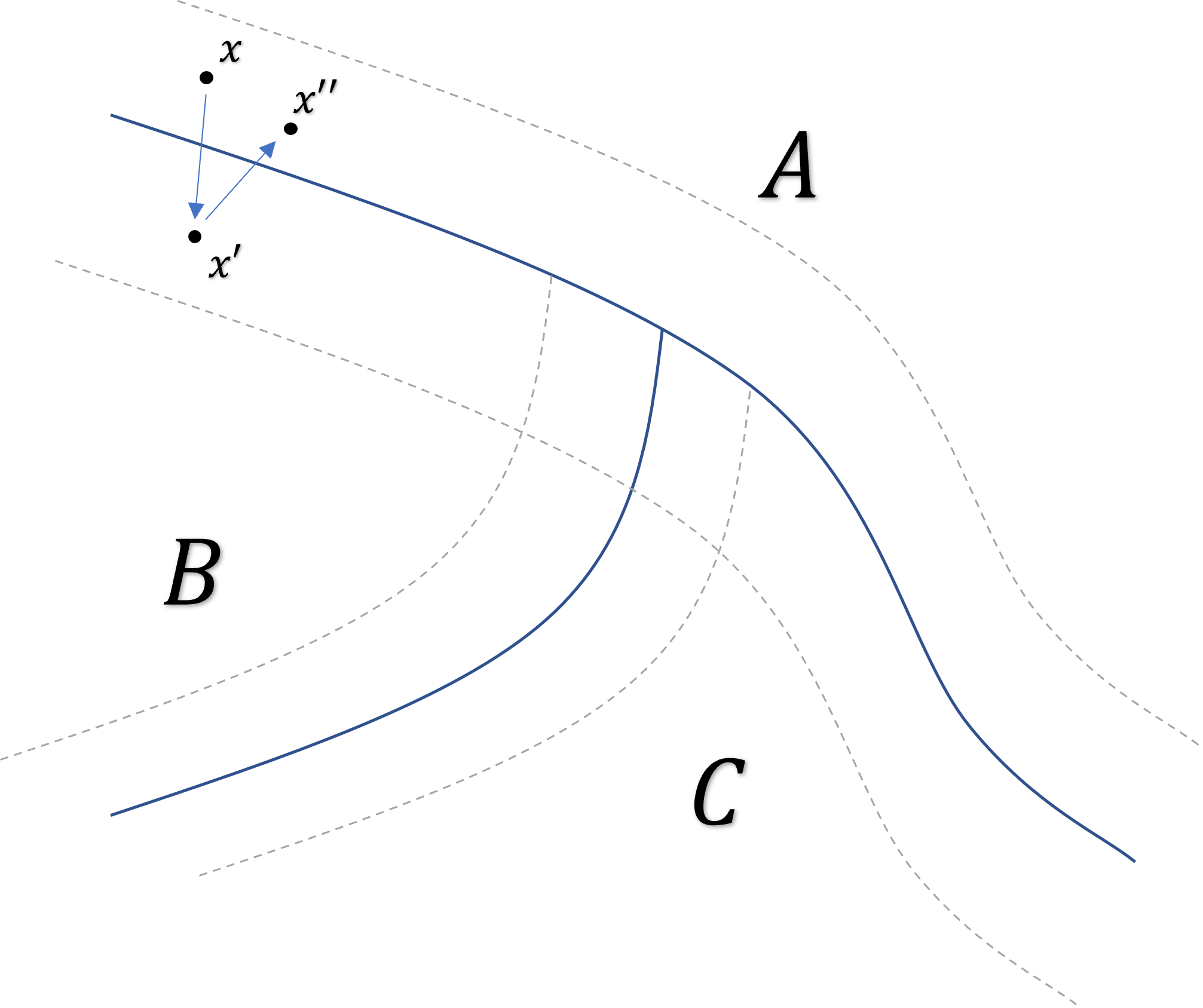}}
\caption{When performing another untargeted adversarial attack on adversarial input, it will cross the decision boundary again and will likely be back to its original class}
\label{fig:neighbor_intuition}
\end{center}
\vskip -0.2in
\end{figure}

We can consider the decision boundaries of a classifier separating three classes A, B and C as shown in \textbf{Figure} \ref{fig:neighbor_intuition}. A natural input x from class A in the confusing region between A and B can be used to create a adversarial input $x^\prime$ under untargeted attack. This input $x^\prime$ is then submitted to the classifier and will be wrongly classified as class B by the given classifier. However, if we perform another untargeted attack on $x^\prime$ with label class B, it will cross the decision boundary again and produce an input $x^\prime$ with class output A. This is the same class as the natural input $x$. The defender does not know whether $x^\prime$ is before or after the attack, but it knows either $x^\prime$ or $x^{\prime\prime}$ will have a high chance of being classified as the same class as the natural input $x$. In other words, $P\left(y=y^\prime\vee y=y^{\prime\prime}\right)$ is high. 

This relation might not hold true when the natural input $x$ is in the confusing region of more than two classes. It will be discussed in \textbf{Section} \ref{discussion}. 

\subsection{Post Training Process}
Once we have found $y^\prime$ and $y^{\prime\prime}$, we know that the ground truth y is highly likely to be one of them. This means we have transformed a multiclass classification task to a simpler binary classification task.  We do this transformation because it is easier to find a robust classifier for a simpler task. We then randomly sample an equal amount of training data from these two classes $y^\prime$ and $y^{\prime\prime}$ respectively.  The sampling is from the same training dataset we used in training stage i.e. there is no need to use additional data. We use these data to do a few iterations of adversarial training to enhance the robustness of our model specifically between these two classes of data. 

The complete algorithm is described in \textbf{Algorithm} \ref{alg:pt_full}. For post training the model with adversarial input generated, we propose two choices of adversarial training: a) ordinary adversarial training, b) fixed adversarial training.

\begin{algorithm}[tb]
   \caption{Post train robust model at inference stage}
   \label{alg:pt_full}
\begin{algorithmic}
   \STATE {\bfseries Input:} Robust model $f_\theta$ with parameter $\theta$, any adversarial input $x^\prime$, training dataset split into all $N$ label classes $D_1…D_N$, post train iteration count $K$, post train batch size $M$
   \STATE {\bfseries Output:} More robust model $f_\theta$ specific to input $x^\prime$
   \STATE Find $y^{\prime\prime}$ using Algorihthm \ref{alg:neighbor}
   \IF{$y^{\prime\prime}==y^{\prime}$}
   \STATE return $f_\theta$       // skip post train without a valid neighbor found
   \ENDIF
   \STATE Sample $M/2$ data from $D_(y^\prime)$ and $D_(y^{\prime\prime})$ respectively
   \FOR{$iter=1$ {\bfseries to} $K$}
   \FOR{$i=1$ {\bfseries to} $M$}
   \STATE from $x_{i}$ generate adversarial input $x_{i}^\prime$
   \STATE $\theta=\theta-\nabla_{\theta} \mathcal{L}\left(f_{\theta}\left(x_{i}^{\prime}\right), y_{i}\right)$ //update using SGD
   \ENDFOR
   \ENDFOR
   \STATE return $f_\theta$
\end{algorithmic}
\end{algorithm}

\subsubsection{Ordinary Adversarial Training}
We can post train the model by ordinary adversarial training with the data from only the original class $y^\prime$ and neighbor class $y^{\prime\prime}$. It is equivalent to solving the following optimization problem:

\begin{align}
\begin{aligned}
\min _{\theta}&\left(\mathbb{E}_{\left(x, y^{\prime}\right) \sim D_{y^{\prime}}}\left\{\max _{\delta \in \Delta} \mathcal{L}\left(f_{\theta}(x+\delta), y^{\prime}\right)\right\} \right. \\
&\left.+\mathbb{E}_{\left(x, y^{\prime \prime}\right) \sim D_{y^{\prime \prime}}}\left\{\max _{\delta \in \Delta} \mathcal{L}\left(f_{\theta}(x+\delta), y^{\prime \prime}\right)\right\}\right)
\end{aligned}
\end{align}

where $D_{y^\prime}$ and $D_{y^{\prime\prime}}$ are distributions of training data with label $y^\prime$ and $y^{\prime\prime}$ respectively.

We found that it is usually better for model trained with certain adversarial training algorithm to continue to be post trained with the same algorithm. For example, models trained with fast-FGSM \cite{wong2020fast} algorithm should be post trained using the Fast-FGSM, while models trained with PGD adversarial training  \cite{madry2019deep} should be post trained using PGD adversarial training.

\subsubsection{Fixed Adversarial Training}
Performing full adversarial training during post training is slow due to the repeated back propagation to find the adversarial input.  We then aim to simplify the adversarial target by fixing the adversarial perturbation to the difference between input $x^\prime$ and its neighbor input $x^{\prime\prime}$.
More specifically, we define this fixed perturbation to be:

\begin{align}
\delta_{f i x}=x^{\prime \prime}-x^{\prime}=\underset{\delta^{\prime} \in \Delta}{\operatorname{argmax}} \mathcal{L}\left(f_{\theta}\left(x^{\prime}+\delta^{\prime}\right), y^{\prime}\right)
\end{align}

We then optimize our model to be robust to the fixed perturbation only by solving:
\begin{align}
\begin{aligned}
\min _{\theta}&\left(\mathbb{E}_{\left(x, y^{\prime}\right) \sim D_{y^{\prime}}}\left\{\mathcal{L}\left(f_{\theta}\left(x+\delta_{f i x}\right), y^{\prime}\right)\right\}\right. \\
&\left.+\mathbb{E}_{\left(x, y^{\prime \prime}\right) \sim D_{y^{\prime \prime}}}\left\{\mathcal{L}\left(f_{\theta}\left(x+\delta_{f i x}\right), y^{\prime \prime}\right)\right\}\right)
\end{aligned}
\end{align}

In other words, the generated adversarial input of $x_{i}$ will be:

\begin{align}
x_{i}^\prime = x_{i} + \delta_{f i x}
\end{align}

Instead of training our model to be robust to all adversarial attacks of data from class $y^\prime$ and $y^{\prime\prime}$, we only post train our model to be robust to any attack similar to the fixed perturbation. Without calculating the adversarial perturbation repeatedly, training using fixed adversarial training is significantly faster. This task is also more specific, hence easier to converge with limited data used during the fine tuning.

\section{Experiments} 
\subsection{Base Model Setups}
All experiments\footnote{All experiment code is available in \url{https://github.com/JokerYan/post_training.git}} are conducted by applying post training techniques on pretrained models in inference stage. The following pretrained models are used as base models:
\begin{enumerate}
\item Fast FGSM \cite{wong2020fast}: pretrained model is provided in the official github\footnote{\url{https://github.com/locuslab/fast_adversarial}} released
\item Madry Model \cite{madry2019deep}: pretrained model is a PyTorch implementation\footnote{\url{https://github.com/louis2889184/pytorch-adversarial-training}} of the original code with verified performance. CIFAR-10 base model is included in the repository and MNIST base model is reproduced.
\end{enumerate}

\subsection{Dataset Related Setups}
Both CIFAR-10 and MNIST datasets are used for evaluation. Setups for both datasets are mostly the same, except for the attack algorithm. To compare the performance with robust accuracy of other published models, the models for the two datasets are attacked with the following setups:
\begin{enumerate}
\item 	CIFAR-10: 20-step $l_{\infty}$  PGD without restart, with $\epsilon=8/255$ and step size $\alpha=3/255$, unless otherwise specified
\item 	MNIST: 40-step $l_{\infty}$  PGD without restart, with $\epsilon=0.3$ and step size $\alpha=0.01$, unless otherwise specified
\end{enumerate}

\subsection{Post Training Setups}
Unless otherwise specified, the model is post trained for 50 epochs with SGD optimizer of 0.001 and momentum of 0.9. The post training inputs are drawn from the original training dataset without additional data. As each model is pretrained with different batch size, the post training batch size also differs. A base model pretrained with batch size of 128, for example, is post trained with 128 inputs from the original class and 128 inputs from the neighbor class, or in other words, a mixed batch of size 256.

Different post training algorithm are used as discussed in \textbf{Section} \ref{post_train_diff_algo}. Fast-FGSM training are marked as “Post Train (fast)”, PGD adversarial training are marked as “Post Train (pgd)”, and fixed adversarial training are marked as “Post Train (fixed)”. In ablation study, normal training instead of adversarial training is also used and is marked as “Post Train (normal)”

\subsection{Evaluation Metrics}
Adversarial accuracy and natural accuracy are used to evaluate the performance of the models with and without adversarial attacks. A more  adversarially robust model should improve the adversarial accuracy while maintaining a good natural accuracy. 

During the evaluation, inputs are provided one by one with test batch size of 1, so that the neighbor class can be found properly. The same adversarial and natural inputs are submitted to the models before and after the post training, to obtain the adversarial and natural accuracy for base and post trained models for comparison. 

\subsection{Robust Accuracy Experiment Results}
\subsubsection{CIFAR-10}
The experiment results on CIFAR-10 are shown in \textbf{Table} \ref{table:robust_acc_cifar}, including the performance of the post trained models, base models, and results published in related works. 

The robust accuracy of Fast FGSM and Madry models can be significantly improved with post train while maintaining the similar natural accuracy. The robust accuracy of Fast FGSM model with post train is also higher than many other white-box defense algorithm, including the UAT++ algorithm which uses 80 million extra unlabeled data.

\begin{table*}[t]
\caption{Robust accuracy of base models and post trained models on CIFAR-10 dataset, compared to performance from other published works}
\label{table:robust_acc_cifar}
\vskip 0.15in
\begin{center}
\begin{small}
\begin{sc}
\begin{tabular}{lcccr}
\toprule
Defense Model & Robust Accuracy & Natural Accuracy \\

\midrule
TRADES \cite{zhang2019theoretically} 		& 0.5661 & 0.8492 \\
\cite{kurakin2017adversarial}       		& 0.4589 & 0.8525 \\
UAT++ \cite{uesato2019labels} with 80M unsup. data 		& 0.6365 & 0.8646 \\

\midrule
Fast FGSM \cite{wong2020fast}  & 0.4681 & 0.8380 \\
Fast FGSM + Post Train (fast)     & 0.6127 & 0.8244 \\
Fast FGSM + Post Train (fix adv) & \textbf{0.6448} & 0.8556 \\
Madry \cite{madry2019deep}                            & 0.4740 & 0.8729 \\
Madry + Post Train (fast)         & \textbf{0.5490} & 0.8750 \\
Madry + Post Train (fix adv)     & 0.5317 & 0.8577
\end{tabular}
\end{sc}
\end{small}
\end{center}
\vskip -0.1in
\end{table*}

\subsubsection{MNIST}
The experiment results on MNIST are shown in \textbf{Table} \ref{table:robust_acc_mnist}, including the performance of the post trained models, base models, and results published in related works. 

The robust accuracy for both Fast FGSM and Madry models are improved from the base model while maintaining a similar natural accuracy. Compared to the results from the related publishes, the post trained models can achieve a comparable robust accuracy with similar natural accuracy, given a good base model. Post trained models on better base model could potentially be even more robust. 

\begin{table*}[t]
\caption{Robust accuracy of base models and post trained models on MNIST dataset, compared to performance from other published works}
\label{table:robust_acc_mnist}
\vskip 0.15in
\begin{center}
\begin{small}
\begin{sc}
\begin{tabular}{lcccr}
\toprule
Defense Model & Robust Accuracy & Natural Accuracy \\

\midrule
TRADES \cite{zhang2019theoretically}	& 0.9607          & 0.9948           \\
Madry (reproduced in \cite{zhang2019theoretically})     & 0.9601          & 0.9936           \\
YOPO	\cite{zhang2019you}	& 0.9627     		& 0.9946                 \\

\midrule
Fast FGSM \cite{wong2020fast} 	& 0.9204          	& 0.985            \\
Fast FGSM + Post Train (fast)  	& 0.9413          	& 0.9847           \\
Fast FGSM + Post Train (fixed) 	& 0.9441          	& 0.9846           \\
Madry \cite{madry2019deep}     	& 0.9536          	& 0.9903           \\
Madry + Post Train (fast)      	& 0.9430           	& 0.9913           \\
Madry + Post Train (fixed)     	& 0.9433          	& 0.9919           \\
Madry + Post Train (pgd)       	& \textbf{0.9618} 	& 0.9874          
\end{tabular}
\end{sc}
\end{small}
\end{center}
\vskip -0.1in
\end{table*}

\subsection{Ablation Study Results}
To analyze the contribution to the performance improvement provided by post training with original and neighbor classes, another set of ablation experiment is conducted to compare the performance of model post trained with different data, on CIFAR-10 dataset.

The first experiment of “50\% original + 50\% random” uses 128 images from the original class and 128 images from a random other class to post train for each batch. The second experiment of “train dataset” uses 256 images randomly sampled from training dataset similar to the normal training process. They are compared to the baseline of “50\% original + 50\% neighbor” which uses 128 images from original class and 128 images from neighbor class. The experiment results are shown in \textbf{Table} \ref{table:ablation}. 

Both models that are post trained without the information of the neighbor class still improve the model without post train, but not as significant as the model post trained with original class data and neighbor class data. This verified our claim that both changing the model and simplifying the task to binary classification problem contribute to a higher robust accuracy.  

\begin{table*}[t]
\caption{Robust accuracy of models post trained with different data compositions}
\label{table:ablation}
\vskip 0.15in
\begin{center}
\begin{small}
\begin{sc}
\begin{tabular}{lccr}
\toprule
Defense Model & Post Train Data & Robust Accuracy \\
\midrule
Fast FGSM \cite{wong2020fast}                   & N.A.                          & 0.4681 \\
Fast FGSM + Post Train (fast) & 50\% original + 50\% random   & 0.5198 \\
Fast FGSM + Post Train (fast) & train dataset                 & 0.579  \\
Fast FGSM + Post Train (fast) & 50\% original + 50\% neighbor & \textbf{0.6448}
\end{tabular}
\end{sc}
\end{small}
\end{center}
\vskip -0.1in
\end{table*}

\subsection{Neighbor Accuracy Experiments}
\label{neighbor_acc_exp}
Since the performance improvement relies on finding the neighbor reliably, we measured the accuracy of finding the correct neighbor. More specifically, a neighbor class $y^{\prime\prime}$ found from original class $y^\prime$ is correct if for the ground truth class $y$, $y=y^\prime$ or $y=y^{\prime\prime}$. 

The experiment is conducted on the entire CIFAR-10 test set, with both untargeted and targeted attacks. In untargeted setups, we apply the untargeted 20 step PGD without restart, with $\epsilon=8/255$ and step size of $3/255$ on the adversarial input provided by the attacker. In targeted setups, we apply the targeted PGD attack with the same parameters as the untargeted attack, but setting the target to all other 9 classes different from the original class.  The class with highest confidence output is selected as the neighbor class. The results are shown in \textbf{Table} \ref{table:neighbor_acc},  where both targeted and untargeted method find the correct neighbor with high accuracy around 87\%.

\begin{table}[t]
\caption{Accuracy of neighbors found using untargeted and targeted attacks}
\label{table:neighbor_acc}
\vskip 0.15in
\begin{center}
\begin{small}
\begin{sc}
\begin{tabular}{lccr}
\toprule
Neighbor Search Method & Neighbor Accuracy \\
 & $P\left(y=y^\prime o r\ y=y^{\prime\prime}\right)$ \\
\midrule
Untargeted             & 0.8707            \\
Targeted               & 0.8783  
\end{tabular}
\end{sc}
\end{small}
\end{center}
\vskip -0.1in
\end{table}

\subsection{Inference Speed Experiments}
Post training algorithm will lead to significantly slower inference stage because of the extra adaptive training involved for each input. The average speed of inference per image is measured on CIFAR-10 test dataset. The experiment is conducted on a single Quadro RTX 6000 gpu. The results are shown in \textbf{Table} \ref{table:inf_speed}.

The experiment result shows that post training improves adversarial robustness at the cost of inference speed, but not every application requires both high robustness and high speed. For example, important government systems may require the user to upload a photo of their relevant ID card. A malicious user could simply print out fake information on a blank piece of paper and add adversarial perturbation to deceive the system that this is indeed the ID type required, a driving license for instance. Such applications may not be very frequently accessed and an inference delay of 10 seconds is tolerable for better secure defense. In such scenarios where inference time of a few seconds is acceptable but high adversarial robustness is needed, post training could be very useful.

\begin{table}[t]
\caption{Inference speed of different post training algorithms}
\label{table:inf_speed}
\vskip 0.15in
\begin{center}
\begin{small}
\begin{sc}
\begin{tabular}{lccr}
\toprule
Model                          & Inference Time \\
			& (second/image) \\	
\midrule
Fast FGSM                      & 0.00173                       \\
Fast FGSM + Post Train (fast)  & 12.78                         \\
Fast FGSM + Post Train (fixed) & 7.75  
\end{tabular}
\end{sc}
\end{small}
\end{center}
\vskip -0.1in
\end{table}

\section{Defense Against Blackbox Attack Using Gradient Estimation}
In addition to the white-box adversarial robustness of our post trained models, we would like to discuss the potential of their robustness against black-box attacks. One important type of blackbox adversarial attack is based on gradient estimation using queries to the model. Chen \cite{chen2017proceedings} proposed doing gradient estimation using Zeroth Order Optimization (ZOO) assuming the attacker can query the probability output (soft label) of each class from the model. Cheng \cite{cheng2018queryefficient} proposed Opt-Attack by estimating the gradient with only the hard labels returned by the model. Others \cite{ilyas2018blackbox, Bhagoji_2018_ECCV, du2018query, tu2020autozoom, ilyas2019prior} also proposed various methods of blackbox attack by estimating gradient with queries to the model. 

With our fine tuning at inference stage, however, the model used in each query will be different. Both the gradient and the probability output used to estimate the gradient will be different during each query. This makes estimating the gradient from both hard and soft label output more difficult resulting in a model that is more robust when facing this type of attack.

Unfortunately, due to the limited speed of fine tuning at inference stage and the large number of queries needed by these attack algorithms, evaluating the defense performance against these attacks directly will take unrealistic amount of time. We have to evaluate the consistency of the actual gradient and gradient estimated to indirectly evaluate the robustness of the model against these attacks.

More specifically, we calculate the gradient of the loss against input pixels of 100 images \footnote{Actual gradient and opt-attack gradient estimated can be calculated with respect to all input pixels of an image at once, so we include all of them in our calculation. ZOO gradient estimation can only calculate the gradient with respect to one pixel value each round, we randomly pick a pixel for each image for calculation.} with a) actual gradient from pytorch b) ZOO estimation \cite{chen2017proceedings} c) Opt-Attack gradient Estimation \cite{cheng2018queryefficient}. These images are the first 100 images from Cifar-10 test set with which the model can actually be fine-tuned (neighbor class different from original class). For each image, we calculate the gradient twice and compare the result by checking whether the gradients have the same sign. 

The gradient and estimated gradient are calculated with the following formula, as described in the respective papers:
Actual Gradient (where $L$ is the loss function and $x_i$ is the input pixel value):
\begin{align}
g=\frac{\partial L(f(x), y)}{\partial x_{i}}
\end{align}

ZOO Gradient (where $h$ is a small constant (set to be 0.0001), $e_i$ represents error vector with only the $i_{th}$ term set to be 1 and the rest to be 0):
\begin{align}
g=\frac{f\left(x+h e_{i}\right)-f\left(x-h e_{i}\right)}{2 h}
\end{align}

Opt-Attack Gradient Estimation (where $v$ is the initial direction, u is a zero-mean gaussian vector, $\beta$ is a small constant setting to be 0.005, $g_{local}$ is the local gradient estimation calculated with binary search at the decision boundary. The detailed algorithm is described in Algorithm 1 from \cite{cheng2018queryefficient} ):
\begin{align}
g=\frac{g_{\text {local}}(v+\beta u)-g_{\text {local}}(v)}{\beta} u
\end{align}

The experiment result of the chance of gradient calculated for the post trained model to have different sign is shown in the \textbf{Table} \ref{table:black_gradient}. $g_1$ and $g_2$ are gradient calculated or estimated for the same model before post training, with the same input.

\begin{table}[t]
\caption{Consistency of gradients calculated/estimated on the post trained model}
\label{table:black_gradient}
\vskip 0.15in
\begin{center}
\begin{small}
\begin{sc}
\begin{tabular}{lccr}
\toprule
Gradient & $P(sign(g_1)\neq sign(g_2))$ \\
\midrule
True Gradient                  & 0.0873 \\
ZOO Gradient Esti.        & 0.46   \\
Opt-Attack Gradient Esti. & 0.58 
\end{tabular}
\end{sc}
\end{small}
\end{center}
\vskip -0.1in
\end{table}

The experiment results show that actual gradient of the model keeps changing each time a query is made. Even with the complete knowledge of the model used for one query, there is a 8.73\% chance to produce a gradient with wrong sign with respect to a pixel, making future attack difficult.
The gradient estimation methods without the knowledge of the model parameters will induce an even larger error when predicting the sign of the gradient. The chance of the ZOO gradient estimation or Opt-Attack gradient estimation to predict a correct sign of gradient is similar to a blind guess, not to mention the actual value of the gradient. This will make the blackbox attack based on these gradients estimated very difficult.

\section{Advantages and Limitations of Post Training a Model}
\label {discussion}
One of the most important advantages of the post training algorithm is the improved robustness on adversarially trained models provided against white-box attacks. By utilizing the randomness in post training as well as the simplified binary classification task between original and neighbor classes, a post trained model can significantly outperform the base model when facing white-box adversarial attacks. 

Another advantage is that this improved robustness is provided without going through the training process of base model again. This is especially important for adversarially trained models as adversarial training is particularly time consuming. Post training algorithm can be easily applied to existing adversarially trained models.

However, post training the model every time during an inference introduces a significant slow down in the inference stage. Although post training a model requires very little data and iterations compared to a full standard training, it is much slower than a normal inference. This greatly limits the usefulness of the post training algorithm in scenarios requiring fast inference.

\begin{figure}[ht]
\begin{center}
\centerline{\includegraphics[width=0.7\columnwidth]{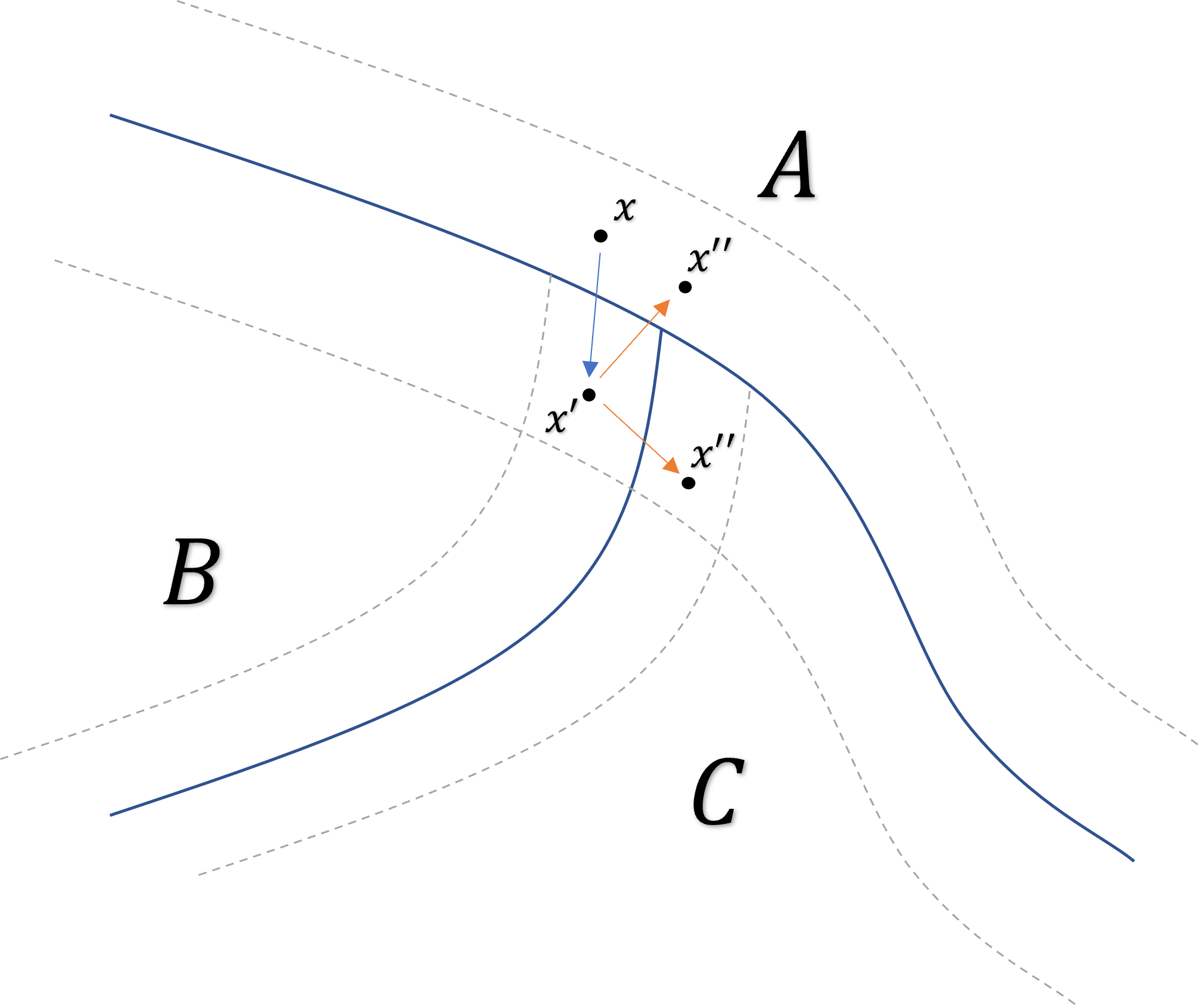}}
\caption{When the input is in confusing region between three or more classes, neighbor found might not be accurate}
\label{fig:neighbor_intuition_three}
\end{center}
\vskip -0.2in
\end{figure}

Another limitation of this algorithm is finding the correct neighbor of an input in the confusing region of more than two classes.  As shown in \textbf{Figure} \ref{fig:neighbor_intuition_three}, an input $x$ around the intersection of decision boundaries of three or more classes might not generate the neighbor input $x^{\prime\prime}$ correctly. Untargeted attack algorithm only finds an adversarial input with a different class output, but does not guarantee to find this input in certain directions. Thus the neighbor $x^{\prime\prime}$ found from another untargeted attack could be of the same class A as the natural input $x$, or of another class C which is not desired. Intuitively, the chance of a random input point falling into the confusing region of three or more classes is lower than the chance of it falling into the confusing region of only two classes, as shown in \textbf{Figure} \ref{fig:neighbor_intuition_three}, but this depends on the characteristics of specific classifier. In our experiment with the Fast-FGSM CIFAR10 classifier, the algorithm can still achieve a high success rate of 87\% when finding the correct neighbor which indirectly validates this intuition.    

\section{Conclusion}
In this paper, we study the effectiveness of finding neighbor class of adversarial input and using it to post train adversarially trained models to be adaptive to any adversarial input. Post training the base model provides stochasticity to the model, which is helpful in defending both white-box and gradient based black-box attacks. Using original class and neighbor class data simplifies the multi-class classification problem, which helps the model to focus on differentiating the true label between  two classes. Our experiments show the post training algorithm significantly improve the adversarial robustness from its base model while maintaining a good natural accuracy.

\bibliography{ca_report}
\bibliographystyle{icml2020}

\end{document}